\documentclass[letterpaper, 10 pt, conference]{ieeeconf}  

\IEEEoverridecommandlockouts                              

\usepackage{amsmath} 
\usepackage{amssymb}  
\usepackage{xurl}
\usepackage{graphicx}
\usepackage{siunitx}
\usepackage{algpseudocode}
\usepackage{subcaption}
\usepackage{booktabs}
\usepackage{multirow}
\usepackage{dblfloatfix}
\usepackage{algorithm2e}
\usepackage[font={small}]{caption}

\title{\LARGE \bf
Motion Primitives Planning For Center-Articulated Vehicles
}

\author{Jiangpeng Hu, Fan Yang, Fang Nan, and Marco Hutter
\thanks{The authors are with Robotic Systems Lab, ETH Zurich, 8092 Zurich, Switzerland.}
\thanks{This project has received funding from the European Union’s Horizon 2020 research and innovation programme under grant agreement No 101070405.}
}

\begin{document}

\maketitle
\thispagestyle{empty}
\pagestyle{empty}

\begin{abstract}
Autonomous navigation across unstructured terrains, including forests and construction areas, faces unique challenges due to intricate obstacles and the element of the unknown.
Lacking pre-existing maps, these scenarios necessitate a motion planning approach that combines agility with efficiency. Critically, it must also incorporate the robot's kinematic constraints to navigate more effectively through complex environments.
This work introduces a novel planning method for center-articulated vehicles (CAV), leveraging motion primitives within a receding horizon planning framework using onboard sensing.
The approach commences with the offline creation of motion primitives, generated through forward simulations that reflect the distinct kinematic model of center-articulated vehicles.
These primitives undergo evaluation through a heuristic-based scoring function, facilitating the selection of the most suitable path for real-time navigation.
To account for disturbances, we develop a pose-stabilizing controller, tailored to the kinematic specifications of center-articulated vehicles. 
During experiments, our method demonstrates a $67\%$ improvement in SPL (Success Rate weighted by Path Length) performance over existing strategies. Furthermore, its efficacy was validated through real-world experiments conducted with a tree harvester vehicle - SAHA.
\end{abstract}
\vspace{0.25cm}
\section{Introduction}
\label{sec:intro}
Autonomous navigation has seen considerable advancements, with vehicles demonstrating remarkable capabilities in real-time decision-making and obstacle avoidance capabilities~\cite{nahavandi2022}.
However, existing research has mostly focused on autonomous cars in urban environments.
The developed methods cannot be directly used for industrial vehicles in construction sites, mining facilities, or farms and forests.
While skilled human drivers can accurately and efficiently maneuver such vehicles in complex environments, the shortage of skilled operators and the potential risk to human drivers~\cite{spectrumnews2023} in these environments highlights the demand for automation.

The environments where most industrial vehicles operate usually contain obstacles of irregular shapes and are densely populated, demanding an agile approach to motion planning.
In addition, industrial vehicles often have special kinematic designs that benefit specific applications, leading to further challenges in planning and control.
For example, center-articulated vehicles (CAVs), due to their adaptability in rugged terrains~\cite{marshall2008}, are commonly used as wheel-loaders, forested harvesters, and tractors.
Their unique steering mechanism, however, poses distinct navigation challenges in addition to the highly unstructured environment.



Traditional planning methods~\cite{Teng2023}, including sampling-based~\cite{lavalle2001}, search-based~\cite{hart1968}, and optimization-based strategies~\cite{jelavic2021}, face challenges in real-world applications, particularly within unknown environments. Those environments are characterized by frequent and unpredictable changes, making it infeasible to maintain an updated global map which sampling-based methods and grid-search methods primarily rely on. Moreover, the special and complex kinematic models of the vehicles make it difficult to find solutions efficiently through optimization-based methods in real-time operations.
End-to-end learning methods, which directly map sensory observations into trajectories or actions, have gained increasing popularity due to their simplicity.
However, learning-based methods struggle to incorporate strict kinematic constraints into the network, leading to potential issues in path feasibility~\cite{zhou2022review}. 
\begin{figure}[t]
    \centering
    \includegraphics[width=0.49\textwidth]{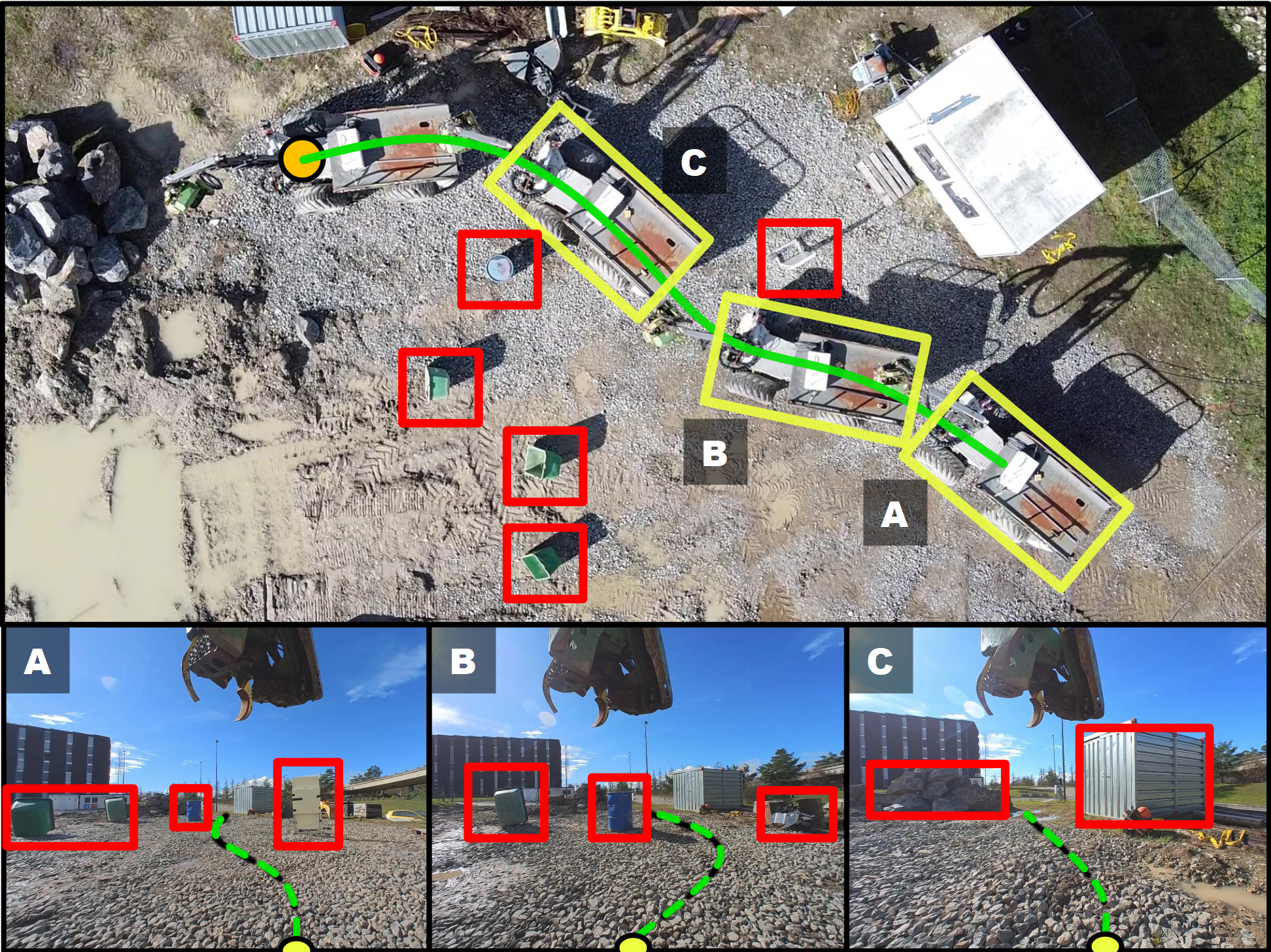}
    \caption{Experiment of a center-articulated vehicle navigating through a construction site. A, B, and C represent three intermediate moments. The goal is set in the orange dot. The green curve shows the vehicle's path avoiding obstacles marked by red boxes. The bottom images illustrate how obstacles influence the planning choices.}
    \label{fig: opening}
    \vspace{-0.5cm}
\end{figure}

Motion primitives-based planning has emerged as a popular approach for autonomous systems, showing significant promise in navigating complex and dynamic environments.
This method, which benefits from the ability to ensure kinodynamic feasibility and efficiency in real-time planning, has been successfully applied across various platforms, from UAVs~\cite{Zhang2020,Dha2020} to AGVs~\cite{Thoresen2021,balkcom2002}.
However, compared to holonomic multi-rotor UAVs, Ackermann steering, and differential wheeled vehicles that have stable and precise motion, center-articulated vehicles present more complex kinematic challenges and restricted maneuverability.
Therefore, existing holonomic primitives for efficient online planning, designed to satisfy geometric continuity and curvature constraints~\cite{Zhang2020}, fall short of capturing the non-holonomic property of CAVs.
This gap highlights the necessity for a dedicated set of primitives taking kinematic constraints and control limitations into account.

This paper addresses the challenge of autonomous navigation for center-articulated vehicles by introducing an algorithm that generates motion primitives tailored to their specific kinematics, combined with a receding horizon strategy for adaptive, real-time planning and a feedback controller for path tracking within kinematic and control constraints.
The main contributions of this paper are summarized as follows:

\begin{enumerate}
    \item A method to generate motion primitives tailored to the kinematics of center-articulated vehicles, which vary with the vehicle's different states, enabling effective real-time planning in a receding horizon strategy.
    \item A pose-stabilizing feedback controller designed to achieve the convergence to a given target under the specific kinematic constraints and control limits of center-articulated vehicles.
    \item Quantitative analysis via simulations in several environments, coupled with robust validation of effectiveness through real-world experiments.
\end{enumerate}
\vspace{0.3cm}

\section{Related Work}
\label{sec:related_work}
Traditional motion planning methods for autonomous driving include sampling-based approaches like the Rapidly-expanding Random Tree (RRT)~\cite{lavalle2001} family (RRT*~\cite{gammell2014}, PRM~\cite{kavraki1996}), grid search methods such as A*~\cite{hart1968} and its variants, and the optimization-based methods~\cite{jelavic2021,guo2008}.
The sampling-based methods and grid search methods have achieved significant strides in structured environments.
For instance, RRT* and hybrid A* have demonstrated success in urban autonomous driving~\cite{hwan2013,montemerlo2008}.
Yet, their application in unstructured environments remains limited due to challenges in adapting to the absence of prior maps.
The optimization-based methods, while allowing for real-time adaptation to dynamic obstacles without the prior map, are not efficient since they solve optimization problems at every step, particularly for vehicles with complex models and environments with numerous obstacles.
In addition, end-to-end planning methods, despite real-time adaptive and efficient, face challenges in accurately integrating kinematic constraints, potentially compromising the feasibility of the generated paths~\cite{chen2023end}.


Motion primitives-based planning, with its capacity to reduce the dispersion of reachable states and planning time, and improve motion plan quality, emerges as a promising solution for autonomous driving.
This approach, by pre-computing and storing motion primitive~\cite{LaValle2006} for quick online reconstruction, enables efficient searching for dynamically feasible trajectories, particularly suited for planning in unknown and unstructured environments~\cite{jarin2021}.
Its applications in trajectory searching have proven successful.
For example, Tobias Loew presents a novel approach to formulating motion primitives as probabilistic distributions of trajectories (ProMP) for AGVs showing higher quality of paths compared to discrete motion primitives~\cite{low2021}.
Marius Thoresen applied motion primitives in the Hybrid A* algorithm for unmanned ground vehicles (UGVs), producing more traversable paths compared with the existing Hybrid A* method~\cite{Thoresen2021}.
In addition, motion primitives are also used in online planning for the high-speed exploration of robots~\cite{Dha2020,Zhang2020}.
In~\cite{Zhang2020}, Zhang and Hu pioneered a method that leverages motion primitives in receding horizon strategy for swift and agile autonomous flight within obstructed environments, proving its effectiveness for drone navigation.
Behind these successful applications, a wide range of methods have been proposed to design motion primitives, which is of fundamental importance for the feasibility and the performance quality of the planning. For example, atomic motion primitives: the well-known Dubins curves~\cite{dubins1957} and Reeds–Shepp curves~\cite{reeds1990}, control sampling primitives and state lattice primitives~\cite{frazzoli2005,pancanti2004,pivtoraiko2011}, and data-driven methods~\cite{kulic2012,deng2018}.
However, related applications to CAVs remain unexplored, with no specialized primitives developed for such vehicles.

The kinematics and dynamics of CAVs have been studied~\cite{DeSantis1997,CorkeRidley2001} to improve their autonomy in challenging environments, yet current planning research primarily leverages traditional methods.
Nayl and Thaker's optimization-based Model Predictive Control (MPC) approach aligns well with kinematics but lacks online efficiency~\cite{Nayl2013}.
Leander Peter's use of Dubins curves achieves path traceability but lacks agility in obstacle-rich environments~\cite{Leander2020}.
For these planning methods, various path-following control approaches were employed: 
Delrobaei and Mehdi's Lyapunov-based approach uses polar coordinates to track the target point but does not inspect the path~\cite{Delrobaei2011},
Altafini proposes a Frenet frame-based path-tracking control~\cite{Altafini1999}, however, the model is complicated.
Ridley and Cork's method~\cite{ridley2003} shows the theoretical feasibility of autonomous regulation but its linearized model is constrained by the assumption of a small steer angle.
A modified pure-pursuit algorithm has been proposed and successfully applied to tractors and rovers offering the advantages of simplicity and stability~\cite{rains2014,fue2020}, yet the steering and speed control are still coupled.
While these studies address specific problems, a systematic planning and control method with efficiency and agility needed for real-time, obstacle-rich environments is yet to be seen on CAVs. 

\vspace{0.3cm}

\section{Methodology}
\label{sec:methodology}
We propose an autonomous navigation system consisting of a local path planner and a path-tracking controller.
The path planner uses motion primitives to generate paths satisfying the kinematic constraints of CAVs, while the tracking controller operating at a higher frequency drives the vehicle along the planned path and accounts for disturbances.
\subsection{Modeling of The Center-Articulated Vehicles}
\label{subsec:kinematic_model}

\begin{figure}[t]
    \centering
    \includegraphics[width=0.42\textwidth]{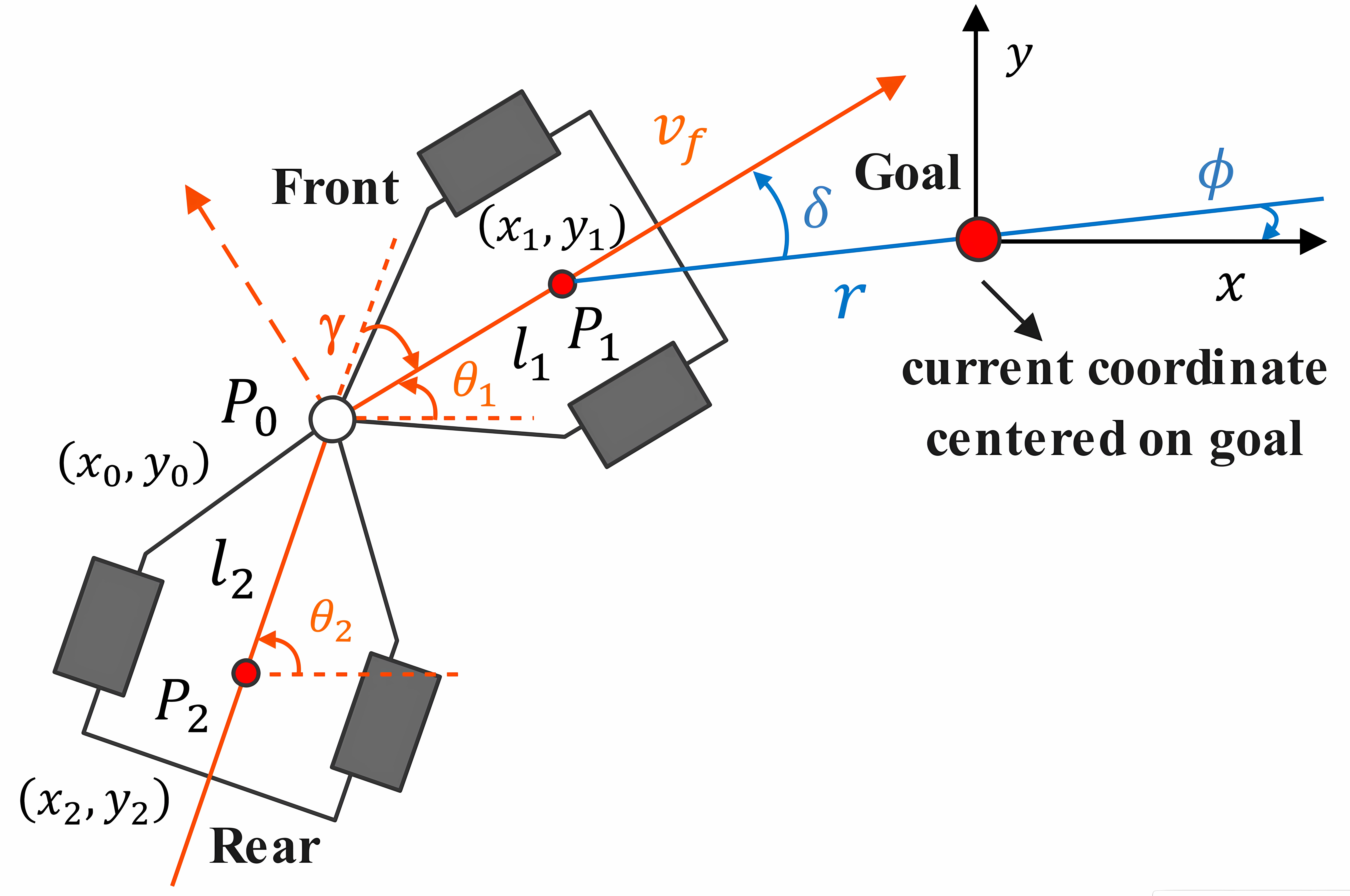}
    \caption{Egocentric polar coordinate system for a CAV during a steady turn}
    \label{fig: steady_turning}
    \vspace{-0.3cm}
\end{figure}

The schematic diagram of a CAV is shown in Fig.~\ref{fig: steady_turning}.
The distances from the central joint to the centers of wheels \( P_1 \) and \( P_2 \) of the front and rear parts are \( l_1 \) and \( l_2 \) respectively.
\( \gamma \) is the center steering angle and the orientation angles of the front and rear parts are \( \theta_1 \) and \( \theta_2 \) respectively.

We define the velocity of the center of the front part \(P_1\) as \(v\).
Assuming no sideslip conditions and considering the geometric constraints at points \( P_1 \) and \( P_2 \), the relationship between the front part's heading, velocity, steering angle, and steering velocity can be expressed as derived in~\cite{CorkeRidley2001}:
\begin{equation}
\dot{x}_1 = v_f \cos(\theta_1) \label{eq:x1_dot}
\end{equation}
\begin{equation}
\dot{y}_1 = v_f \sin(\theta_1) \label{eq:y1_dot}
\end{equation}
\begin{equation}
\dot{\theta}_1 = - \frac{v_f \sin(\gamma) + l_2 \dot{\gamma}}{l_1 \cos(\gamma) + l_2} . \label{eq:theta1_dot}
\end{equation}

By organizing equations \eqref{eq:x1_dot} to \eqref{eq:theta1_dot}, we derive a kinematic model with four state variables, using the front center \(P_1\) as the virtual control point:
\begin{equation}
\begin{bmatrix}
\dot{x}_1 \\
\dot{y}_1 \\
\dot{\theta}_1 \\
\dot{\gamma}
\end{bmatrix} =
\begin{bmatrix}
\cos(\theta_1) & 0 \\
\sin(\theta_1) & 0 \\
-\frac{\sin(\gamma)}{L} & -\frac{l_2}{L} \\
0 & 1
\end{bmatrix}
\begin{bmatrix}
v \\
\dot{\gamma}
\end{bmatrix}, \label{eq:kinematic_model_subset}
\end{equation}
where \( L = l_2 + l_1 \cos(\gamma) \) is defined as the effective length of the vehicle.

\subsection{Offline Primitives Generation}
\label{subsec:section2}
The generation of primitives refers to the generation methods of state lattice primitives and control sampling primitives.
According to the kinematic model in Equation \eqref{eq:kinematic_model_subset}, the vehicle's state space includes \( x_1 \), \( y_1 \), \( \theta_1 \), \( \gamma \), while the control space includes \( v \) and \( \dot{\gamma} \).
First, we discretize the state space to create different groups of primitives.
For the egocentric local planning problem, vehicle-centered coordinate frames can be used, which simplifies the vehicle's state space as \( (x_1, y_1) = (0, 0) \) and \( \theta_1 = 0 \).
Thus, we only discretize \( \gamma \) into 30 groups, serving as the vehicle's state lattices: \(\Sigma_{i}, i \in [1,30] \).

\begin{figure}[t] 
    \centering
    \includegraphics[width=0.48\textwidth]{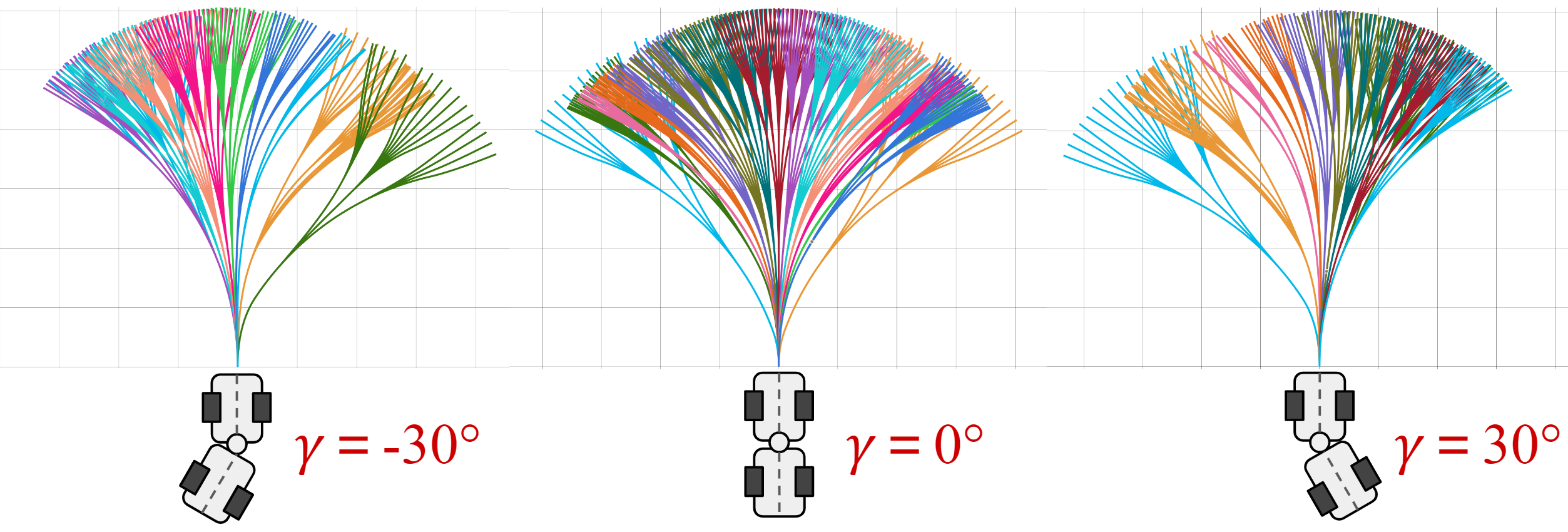}
    \caption{Example of the motion primitives for vehicles with steering angles equal -30 degrees, 0 degrees, and 30 degrees. Each color denotes a control group of trajectories that coincide during the initial period and split twice as time progresses.}
    \label{fig:motion_primitives}
    \vspace{-0.3cm}
\end{figure}

We define 15 control groups inside each state lattice \(\Sigma_i\) as trajectories that vary from different initial control inputs.
Each control group \(\Sigma_{ij}, (j \in [1,15])\) has different initial control inputs designed based on the mechanical limitations of vehicles and excluding significant overlap.
Each control group contains trajectories that start with the same \( v \) and \( \dot{\gamma} \) and split twice at \(t = (3/v_j)\)s and \(t = (6/v_j)\)s.
Trajectories in each control group are generated using forward kinematics simulation:
\begin{equation}
\tau^{ij}_k \in T_i(\gamma_i, v_j, \dot{\gamma}_j) = \left\{ F(\gamma_i, v_j, \dot{\gamma}_j, t \in [0: \Delta t: T_j]) \right\} 
\label{eq:trajectory_generation_adjusted}
\end{equation}
\begin{equation}
\Sigma_i = \{\tau^{ij}_k \mid  j \in [1,15], k \in [1,N_{ij}] \}
\end{equation}
\begin{equation}
\Sigma_i \ni \Sigma_{ij} = \{\tau^{ij}_k \mid  k \in [1,N_{ij}] \}.
\end{equation}
Here, \( \Delta t = 0.1s\), and \( T_j = (10 / v_j) s\), ensuring equal field of view of  \( 10 m \) for trajectories at different speeds.
This horizon length is tuned to balance the computational complexity and prediction accuracy.
\( k \in [1, N_{ij}] \) is the ID of trajectories within each state lattice.
After adjustments and optimization, each state lattice contains an average of 450 trajectories \begin{math}\left(\sum_{j=1}^{15} N_{ij} = 450, i \in [1,30]\right)\end{math}.
Fig.~\ref{fig:motion_primitives} illustrates the primitives for 3 typical state lattices, highlighting the significant impact of initial \( \gamma \) on trajectory formation.

\subsection{Online Receding-horizon Planner}
\label{subsec: online planner}
\begin{figure}[b]
  \centering
  \begin{subfigure}[b]{0.22\textwidth}
    \centering
    \includegraphics[height=2.5cm]{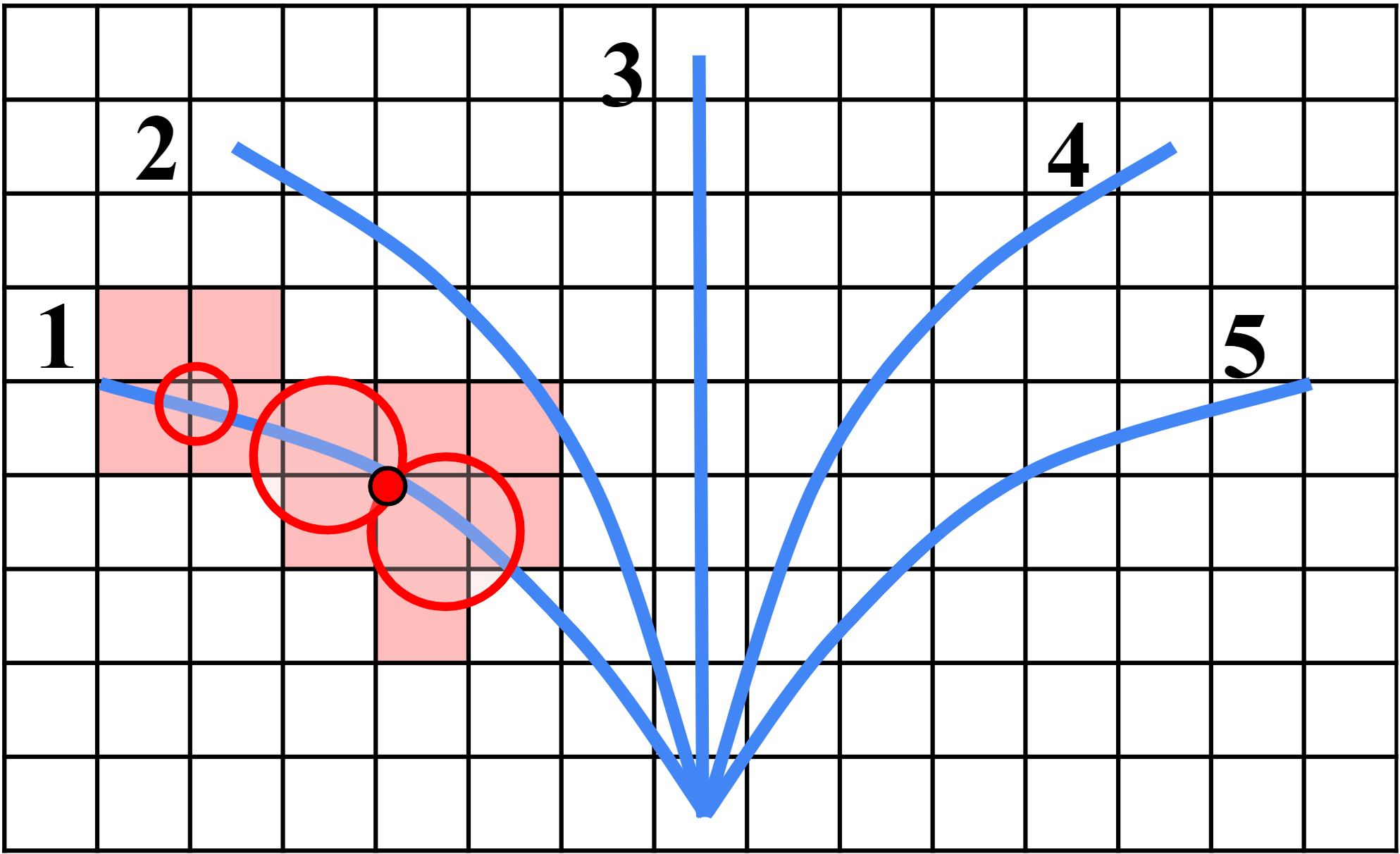} 
    \caption{Offline grid marking}
    \label{fig:coorespondence}
  \end{subfigure}\hfill
  \begin{subfigure}[b]{0.26\textwidth}
    \centering
    \includegraphics[height=2.5cm]{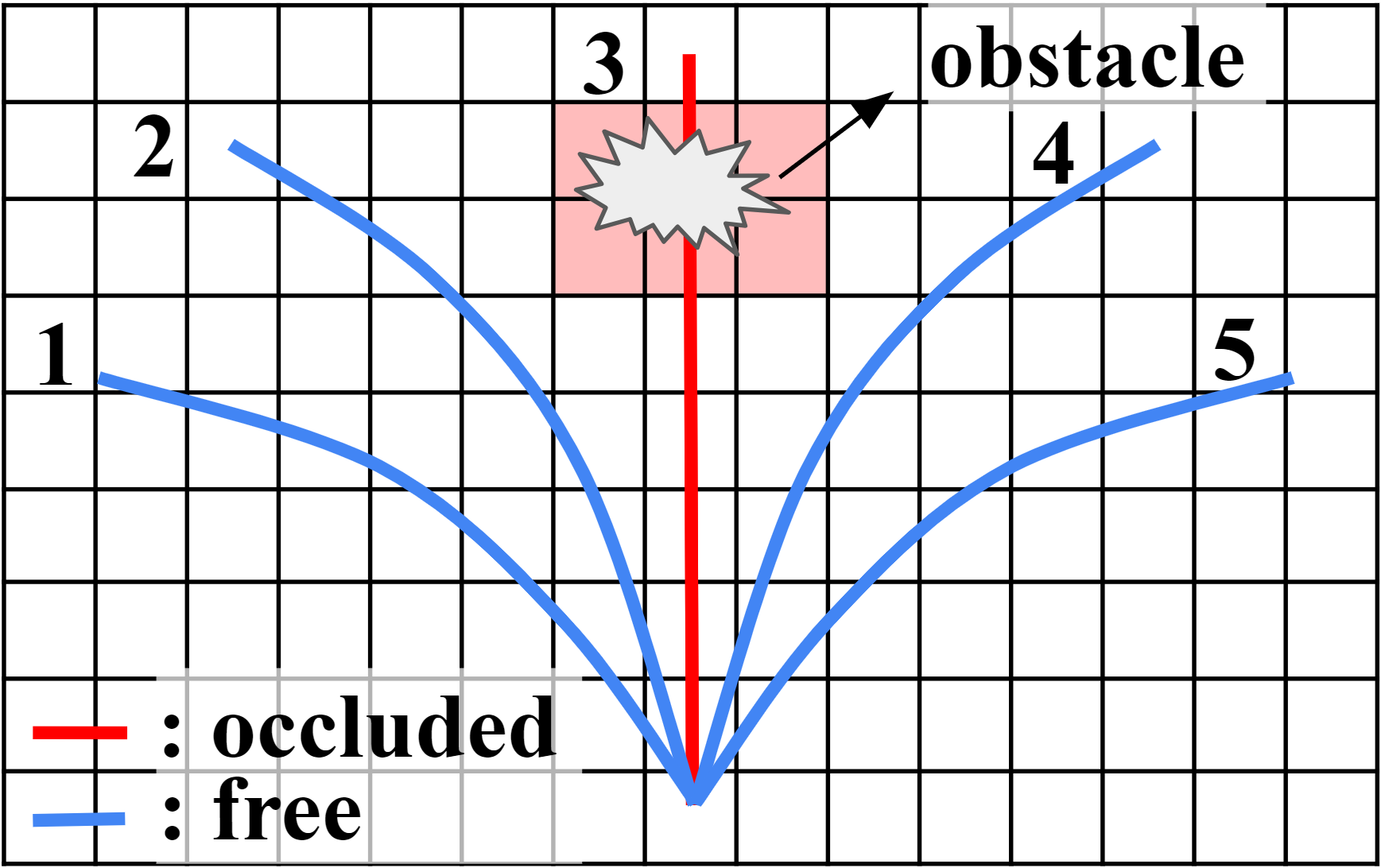} 
    \caption{Online collision detection}
    \label{fig:path_elimination}
  \end{subfigure}
  \caption{An illustration of the two-step collision detection: (a) shows an example of marking the corresponding collision grid for the Trajectory \(1\). The collision body for the CAV is designed as 2 circles around the front and rear parts and an extra circle for the arm. (b) shows an online collision detection where Trajectory \(3\) is marked as occluded.}
  \label{fig:collision_detection}
  \vspace{-0.3cm}
\end{figure}

Primitives set generated in Sec.~\ref{subsec:section2} are further processed for collision detection.
Drawing inspiration from Zhang's method~\cite{Zhang2020}, the proposed two-step collision detection method combines offline precomputation and online path selection.
Initially, potential collision points along trajectories in primitive sets are identified and mapped to grid cells, which are later filled with real-time terrain data.
During online running, the planner dynamically eliminates paths based on real-time obstacle data matched to these grids, ensuring efficient and responsive navigation without excessive computational demands.
As shown in Fig.~\ref{fig:collision_detection}.

We apply a heuristic-based scoring function to evaluate collision-free trajectories in the receding horizon planner.
Given the adoption of a receding horizon control strategy, our focus is primarily on imminent trajectories.
Since trajectories within the same control group coincide during the initial period (see Fig.~\ref{fig:motion_primitives}), we calculate the comprehensive score \(S_{ij}\) for each control group of trajectories instead of each trajectory:
\begin{equation}
S_{ij} = \frac{\sum_{k=1}^{N_{ij}} s^{ij}_k}{N_{ij}} , \label{eq:comprehensive_score}
\end{equation}
where \(N_{ij}\) is the total number of trajectory in control group \(\Sigma_{ij}\).
Each trajectory score \(s^{ij}_k\) is determined by
\begin{equation}
\begin{split}
s^{ij}_k = & (s^{dir} + \alpha \cdot s^{dist})^2 \cdot s^{vel} \cdot s^{state} \cdot s^{terrain} \cdot s^{p}.
\end{split}
\label{eq:trajectory_score}
\end{equation}
The score terms in (\ref{eq:trajectory_score}) are given by:
\begin{align}
    s^{dist} =& f^{dist}\propto(D_{max} - \sqrt{dx^2 + dy^2})\\
    s^{dir} =& f^{dir}\propto{(2\pi-|d\theta}_1|),({2\pi-|d\theta}_2|)\\
    s^{state} =& f^{state}\propto{(2\gamma_{max} - |\gamma_k - \gamma_{now}|)}\\
    s^{vel} =& f^{vel}\propto{v_k}\\
    s^{terrain} =& f^{terrain}(H_{max})\\
    s^{p} =& f^{p}(\mathit{diff_c}),
\end{align}
where \((dx, dy)\) is distance to the goal, \(D_{max}\) is a constant.
\({d\theta}_1\) and \({d\theta}_2\) are direction differences toward the goal based on the heading and position of the trajectory \(\tau^{ij}_k\).
\( \gamma_k \) is the initial steer angle for the trajectory \(\tau^{ij}_k\).
\( v_k \) represents the velocity for the trajectory \(\tau^{ij}_k\).
\(H_{max}\) is the maximum height of the terrain along the trajectory: \begin{math}\max ( height ) \text{ in } \tau^{ij}_k\end{math}.
\begin{math}\mathit{diff_c}\end{math} is the distance between the end position of \(\tau^{ij}_k\) and the end position of the last selected trajectory.

\begin{figure}[t]
  \centering
  \begin{subfigure}{0.22\textwidth}
    \centering
    \includegraphics[height=2.4cm]{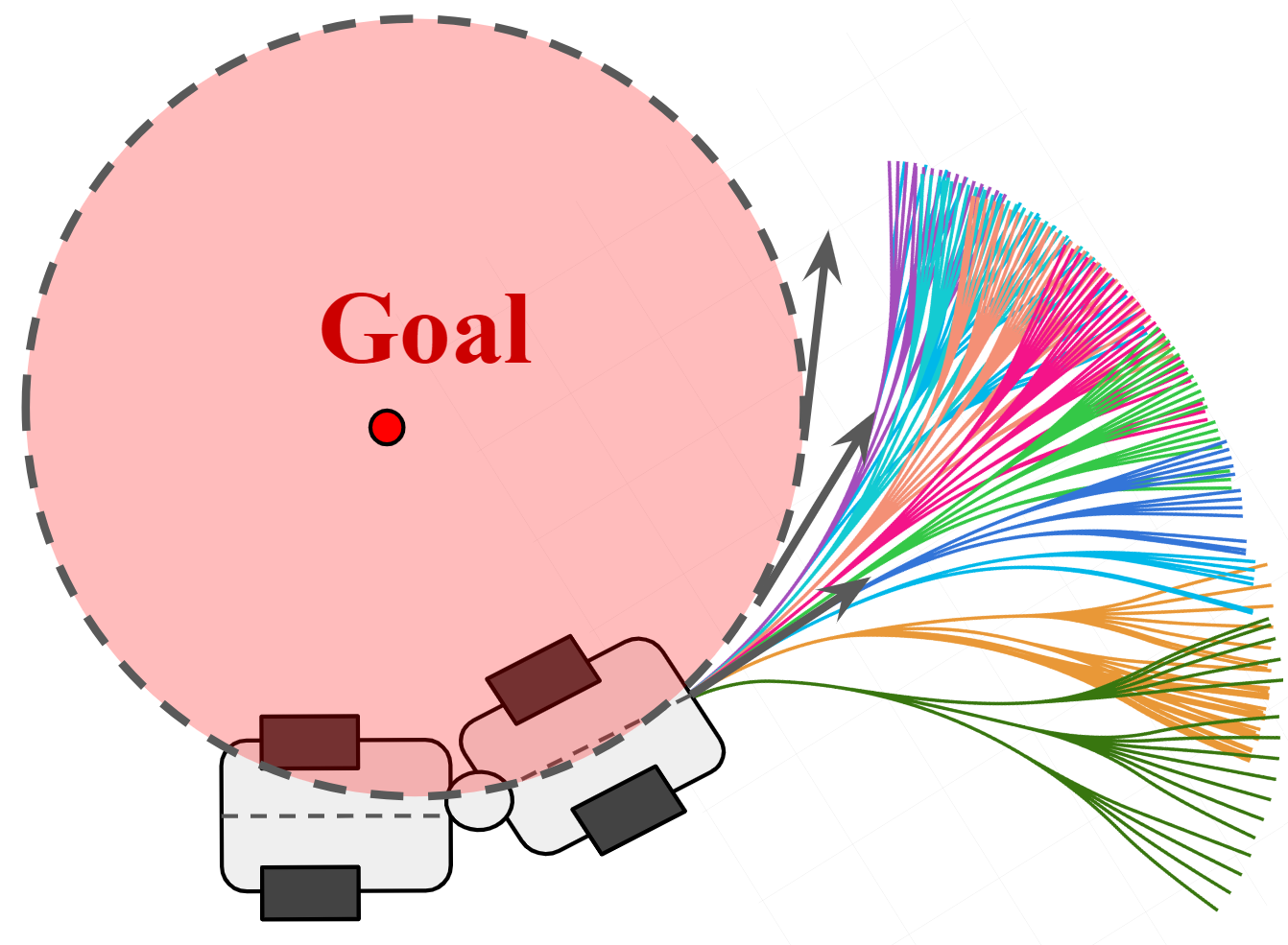} 
    \caption{``Unreachable zone``}
    \label{fig: unreachable zone}
  \end{subfigure}\hfill
  \begin{subfigure}{0.26\textwidth}
    \centering
    \includegraphics[height=2.4cm]{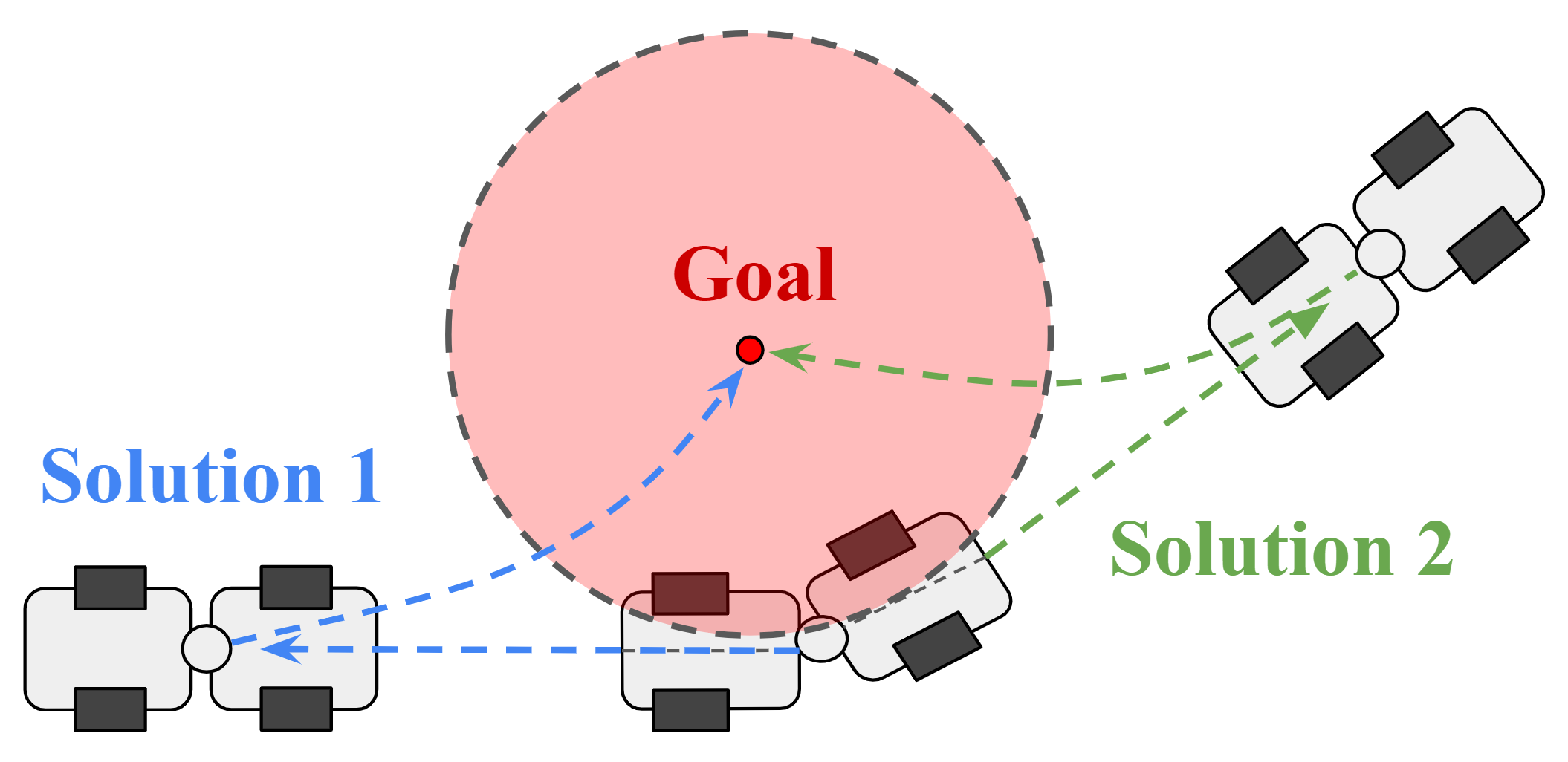} 
    \caption{Bi-directional trajectories}
    \label{fig: special strategy}
  \end{subfigure}
  \caption{(a) shows the ``Unreachable zone`` for the CAVs. (b) depicts two potential solutions with bi-directional trajectories.}
  \label{fig: unreachable zone&special strategy}
  \vspace{-0.3cm}
\end{figure}

Like most non-holonomic vehicles, CAVs have a minimum turning radius due to mechanical constraints.
This creates a circular "unreachable zone" around the vehicle when it reaches its maximum limit, as illustrated in Fig.~\ref{fig: unreachable zone}.
To reach this zone, bi-directional trajectories are required. Our primitives, focusing on continuity and stability, are configured for unidirectional movement (either forward or backward).
This setting delegates the responsibility of initiating direction changes to the online planner by defining a special case in computing the score function.

For goals set within this zone, the search begins at the goal using primitives with local coordinates heading to all directions to find intersections with vehicle-originating trajectories, indicating potential future positions.
We evaluate trajectories heading in the opposite direction from the goal, merging two opposing paths into a geometrically discontinuous route.
The trajectory's overall score is determined by averaging the scores of both segments.
Fig.~\ref{fig: special strategy} demonstrates two such paths to the goal.

Following previous methods~\cite{Zhang2020}, we take the coinciding part before the first split (as discussed in Sec.~\ref{subsec:section2}) of the trajectories in the control group with the highest cumulative score as the optimal path.
To sum up, the online local planning algorithm is described in Algorithm~\ref{alg:online_local_planning}.
\RestyleAlgo{ruled}
\SetAlgoNlRelativeSize{-1}
\SetAlCapFnt{\small}
\SetAlCapNameFnt{\small}
\SetAlFnt{\small}
\begin{algorithm}[ht]
\caption{Online local planning algorithm}\label{alg:online_local_planning}
\DontPrintSemicolon
\SetKwInOut{Input}{input}\SetKwInOut{Output}{output}
\Input{primitives sets, correspondences array, vehicle’s state, sensor data}
\Output{$\tau^*$(best path) or no-path-found}
\Begin{
    Identifies obstacle grids using sensor data; \;
  \For{each obstacle grid}{
    Label correspondent paths as occluded;\;
  }
  \For{each direction}
  {
      \For{each collision-free path}
      {
      \eIf{goal in “Unreachable Zone”}{
        Find intersects paths* towards the goal;\;
        Compute average score $s_{k}$ for path*;\;
        }{
        Compute $s_{k}$ based on score function;\;
        }
      }
      Compute $S_{ij}$ for each control group; \;
  }
  \eIf{All path occluded}
  {
    Return no-path-found; \;
}
    {
    Find max score $S_{ij}$, return the first coinciding part of $\tau^*$ from the control group with highest $S_{ij}$;\;
  }
}
\end{algorithm}

\subsection{Pose-Stabilizing Control Law}
\label{subsec: path tracking}

The motion primitive selected by the planner provides a kinematically feasible path, but it cannot be used as a feedforward control input due to the discretization of states in the primitive generation process.
Therefore, we sample a lookahead point from the selected primitive and design a variant of the pure-pursuit method inspired by Park's method~\cite{Park2011}. 
Compared to the pure-pursuit method, our controller accounts for the specialized kinematics of CAVs.
It starts with the error definition in the polar coordinate.
As shown in Fig.~\ref{fig: steady_turning}, the distance to the goal is \( r \).
Let \( \phi \in (-\pi, \pi] \) represent the orientation of \( P_1 \) relative to the goal.
Let \( \delta \in (-\pi, \pi] \) denote the vehicle's heading orientation relative to the lines toward the goal.

In the polar coordinate system, it can be easily derived that:
\begin{equation}
\begin{bmatrix}
\dot{r} \\
\dot{\phi} \\
\dot{\delta}
\end{bmatrix} =
\begin{bmatrix}
-\cos(\delta) & 0 \\
\frac{\sin(\delta)}{r} & 0 \\
\frac{\sin(\delta)}{r} & 1
\end{bmatrix}
\begin{bmatrix}
v \\
\omega
\end{bmatrix} \label{eq:system_error_matrix}
\end{equation}

Where the angular velocity $\omega$ is given by:
\begin{equation}
\omega = \dot{\theta}_1 = -\frac{\sin(\gamma) v}{l_2 + l_1 \cos(\gamma)} -\frac{l_2 \dot{\gamma}}{l_2 + l_1 \cos(\gamma)} \label{eq:omega}
\end{equation}

Equations~\eqref{eq:system_error_matrix} characterize the geometric error between the current state and the desired goal.
It should be noted that the control variable \(\omega\) is influenced by the actual control variables: the velocity \(v\) and the steering velocity \(\dot{\gamma}\).

Assuming a positive and non-zero velocity \(v\), with \(\omega\) as the sole control signal, \(\omega\) primarily influences the state \(\delta\), while \(r\) and \(\phi\) are governed by \(\delta\).
As \(r\) and \(\phi\) define the vehicle's position and \(\delta\) its steering, we can decouple the system into 2 systems:

\begin{equation}
\label{eq: slow system}
\begin{bmatrix}
\dot{r} \\
\dot{\phi}
\end{bmatrix} =
\begin{bmatrix}
-\cos(\delta) \\
\frac{\sin(\delta)}{r}
\end{bmatrix} v
\quad 
\end{equation}

\begin{equation}
\label{eq: fast system}
\dot{\delta} = \frac{\sin(\delta) v}{r} + \omega
\quad 
\end{equation}

\subsubsection{Slow Subsystem and the Reference Heading}

For the slow subsystem given by equation~\eqref{eq: slow system}, we introduce a virtual control for the heading \(\delta\) as:
\begin{equation}
\delta = \arctan(-k_\phi \phi) \label{eq:virtual_control}
\end{equation}
where \( k_\phi \) is a positive constant.
Following~\eqref{eq:virtual_control}, the slow subsystem~\eqref{eq: slow system} can be proved to be Lyapunov stable.
This virtual control law can be interpreted as the reference heading of the vehicle based on its current state \( \phi \).
It delineates the slow manifold to which the vehicle's fast heading dynamics will eventually converge. 

\subsubsection{Fast Subsystem and Closed-Loop Steering}

The fast subsystem~\eqref{eq: fast system} aims to steer the vehicle to reach the reference heading in the slow subsystem~\eqref{eq: slow system}.
For the virtual control variant \(\omega\), the control law is given by: 
\begin{equation}
\omega = \kappa(r, \phi, \delta) \cdot v
\end{equation}
where \(\kappa\) is the curvature of the path resulting from our proposed control law. The curvature \(\kappa\) can be expressed as:
\begin{equation}
\kappa = -\frac{1}{r} \left[ k_{\delta}(\delta - \tan^{-1}(-k_{\phi} \phi)) + \left(1 + \frac{k_{\phi}}{1 + (k_{\phi} \phi)^2}\right) \sin(\delta) \right]
\end{equation}

Considering the kinematic model for CAVs, we then determine steering velocity \(\dot{\gamma}\) as:
\begin{equation}
\dot{\gamma} = \kappa^*(r, \phi, \delta, \gamma) \cdot v \label{eq: gamma_dot}
\end{equation}
\begin{equation}
\kappa^*(r, \phi, \delta, \gamma) = -\left[ \frac{L}{l_2}\kappa(r, \phi, \delta) + \frac{\sin(\gamma)}{l_2} \right] \label{eq: kappa_dot}
\end{equation}

we designed \(v\) as a function of \(\kappa\):
\begin{equation}
     v = v(\kappa) = \frac{v_{op}}{1 - \beta |\kappa|^\lambda} \label{eq: final_v}
\end{equation}
\begin{equation}
     v_{op} = v_{max}(1 - \beta|\kappa_{max}|^\lambda)
\end{equation}
where, $v_{op}$ is the operational speed and $v_{ax}$ is the maximum linear velocity.
$\beta$ and $\lambda$ are design parameters, chosen such that $v \rightarrow v_{op}$ as $\kappa \rightarrow 0$, and $v \rightarrow v_{max}$ as $\kappa \rightarrow \kappa_{max}$. 

The control law defined by Equations~\eqref{eq: gamma_dot} to \eqref{eq: final_v} fundamentally relies on pre-computed primitives for theoretical stability since both \(\gamma\) and \(\dot{\gamma}\) are mechanically constrained.
When \(\gamma\) or \(\dot{\gamma}\) reaches the limit, Equation~\eqref{eq: gamma_dot} might not be able to satisfy, resulting in potentially slow or even failure in the convergence of slow subsystem.
This issue is addressed by constraining each point in the tracking paths to specifically designed primitive sets. These sets are tailored to the kinematic and mechanical constraints of center-articulated vehicles (CAVs), thereby facilitating system convergence without violating the outlined constraints.
\section{Experiments}
\label{sec:Experiments}
\begin{figure}[b]
    \centering
    \includegraphics[width=0.25\textwidth]{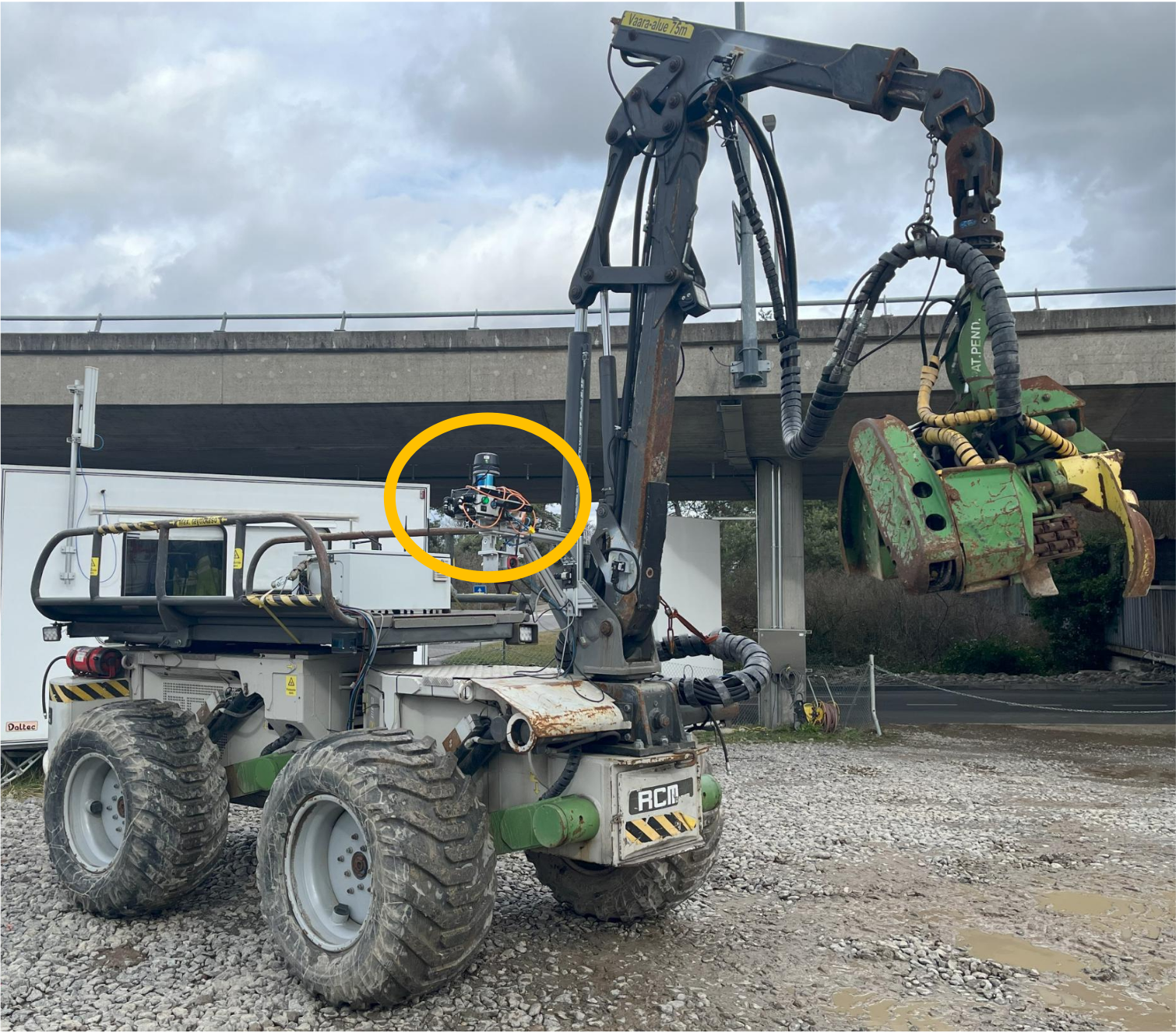}
    \caption{SAHA in a construction site, the yellow circle indicates the vehicle-mounted LiDAR.}
    \label{fig:harveri_real&gazebo}
\end{figure}

We evaluate the proposed method in simulation environments and compare its performance with the baseline before carrying out real-world experiments.
The test platform, SAHA (Supervised Autonomous HArvester)~\cite{Jelavic22HarveriSmall}, is an autonomous tree harvester with a center-articulated chassis.
The vehicle is equipped with a LiDAR for real-time environmental mapping.
Key specifications of SAHA are listed in Table~\ref{tab:harveri_specs}. Fig.~\ref{fig:harveri_real&gazebo} depicts the actual SAHA vehicle.

\begin{table}[t]
\centering
\caption{SAHA Characteristics}
\label{tab:harveri_specs}
\resizebox{0.35\textwidth}{!}{
\renewcommand{\arraystretch}{1} 
\begin{tabular}{ll}
\toprule
\textbf{Specification} & \textbf{Value} \\
\midrule
Vehicle Weight & 4500 kg \\
Chassis Length & 4.22 m \\
Chassis Width & 2.2 m \\
Distance (Front Axle to Hitch, \( l_1 \)) & 0.95 m \\
Distance (Rear Axle to Hitch, \( l_2 \)) & 0.95 m \\
Wheelbase & 1.6 m \\
Max Articulation Angle & 33° \\
\bottomrule
\end{tabular}}
\end{table}

\begin{figure*}[t]
    \centering
    \includegraphics[width=0.99\textwidth]{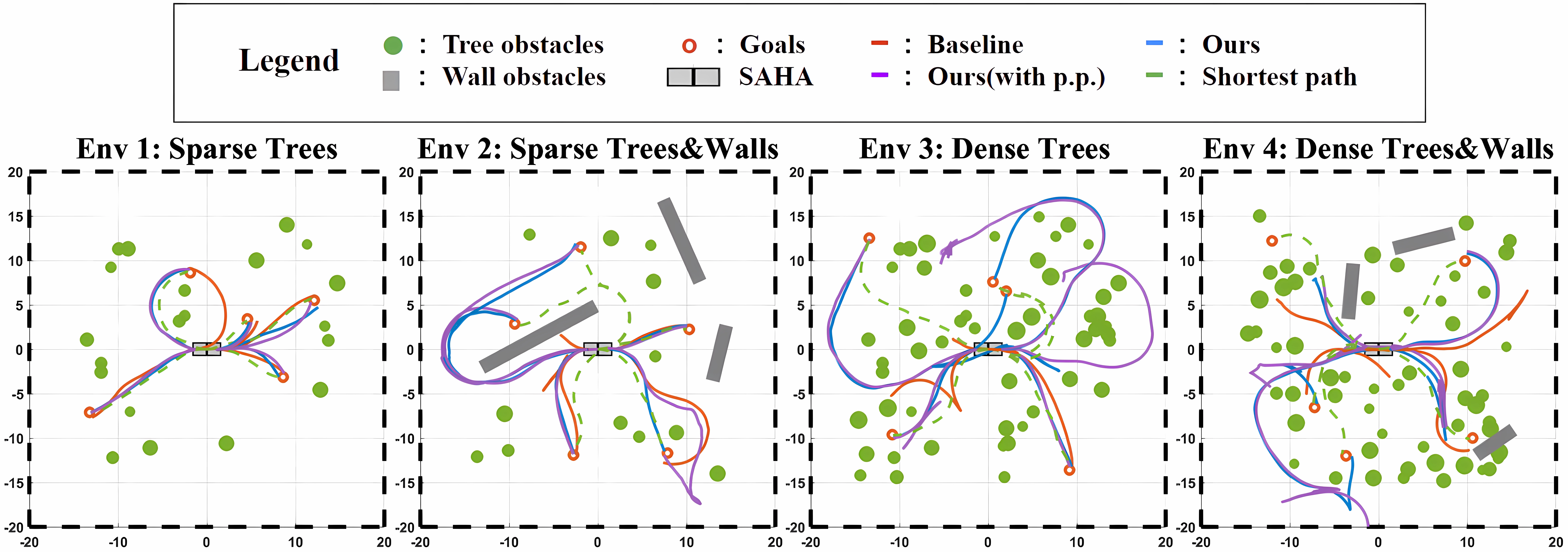}
    \caption{Example maps from the four types of environments: 1) sparse tree obstacles ($N_t=20$), 2) sparse tree and wall obstacles ($N_t=12$, $N_w=3$), 3) dense tree obstacles ($N_t=50$), 4) dense tree and wall obstacles ($N_t=55$, $N_w=3$). On maps there are examples of trajectories using different methods: optimal paths (green), our method (blue), the baseline method (red), and our method with p.p. (purple)}
    \label{fig:trajs}
\end{figure*}

\begin{table*}[t]
\centering
\renewcommand{\thetable}{3} 
\caption{Comparison of Motion Planning Test Results}
\label{tab:test_results}
\resizebox{0.99\textwidth}{!}{%
\begin{tabular}{lcccccccccc}
\toprule
 & \multicolumn{2}{c}{Env 1} & \multicolumn{2}{c}{Env 2} 
 & \multicolumn{2}{c}{Env 3} & \multicolumn{2}{c}{Env 4}
 & \multicolumn{2}{c}{Total}\\
\cmidrule(lr){2-3}
\cmidrule(lr){4-5}
\cmidrule(lr){6-7}
\cmidrule(lr){8-9}
\cmidrule(lr){10-11}
 & SR  & SPL & SR & SPL & SR  & SPL & SR  & SPL & SR  & SPL 
 \\
\midrule
Baseline & 92\% & 0.8047 & 48\%  & 0.4221 & 40\% & 0.3287 & 36\% & 0.3110 & 54\% & 0.4666 \\
Ours (with p.p.) & \textbf{100\%} & \textbf{0.9632} & 96\% & 0.7451 & 92\% & 0.6928 & 64\% & 0.5426 & 88\% & 0.7359 \\
Ours & \textbf{100\%} & 0.9570 & \textbf{100\%} & \textbf{0.7987} & \textbf{96\%} & \textbf{0.7550} & \textbf{76\%} & \textbf{0.5999} & \textbf{93\%} & \textbf{0.7776}\\
\bottomrule
\end{tabular}
}
\end{table*}

In both simulation and real-world experiments, the proposed planning method was implemented with replan frequency at \SI{20}{\hertz}, and the control loop is executed at \SI{50}{\hertz}.
The look-ahead distance for the controller is set as \SI{1.5}{m}, which balances the responsiveness and the smoothness of trajectories. 
The same values were used for the baseline method in the simulation.
For controller, we choose parameters as: \(k_{\phi} = 0.5, k_{\delta} = 1.0, \beta = 0.5, \lambda = 1\).

\subsection{Simulation Experiments}
The simulation experiment of SAHA is performed in the Gazebo simulator.
We first evaluate the proposed path-following controller independently before using it in the autonomous driving pipeline.
The controller is tested against the modified pure-pursuit approach, with the reference paths derived from generated primitives that accommodate different vehicle states.
Specifically, three states are tested, each assessed across 90 trajectories.
Cross-track error (CTE)~\cite{Lekkas2014} is used to evaluate the tracking error.
As shown in Table~\ref{tab:tests_control}, our method's advantage grows with the tracking steering angle \(\gamma\) increases, showing a 4.73\% reduction of CTE at \(\gamma = 0^\circ\) and 27\% at \(\gamma = 30^\circ\).
\begin{table}[t]
\centering
\renewcommand{\thetable}{2} 
\caption{CTE Comparison of Two Controller}
\label{tab:tests_control}
\resizebox{0.48\textwidth}{!}{
\renewcommand{\arraystretch}{1} 
\begin{tabular}{lcccc}
\toprule
& \multicolumn{1}{c}{$\gamma = 0^\circ$} & \multicolumn{1}{c}{$\gamma = 15^\circ$} & \multicolumn{1}{c}{$\gamma = 30^\circ$} & \multicolumn{1}{c}{Total} \\
\midrule
Pure-pursuit & 0.0338  & 0.0424 & 0.0615 & 0.0459 \\
Ours & \textbf{0.0322} &\textbf{ 0.0388} & \textbf{0.0448} & \textbf{0.0386}\\
\bottomrule
\end{tabular}}
\end{table}

\begin{figure}[t]
    \centering
    \includegraphics[width=0.32\textwidth]{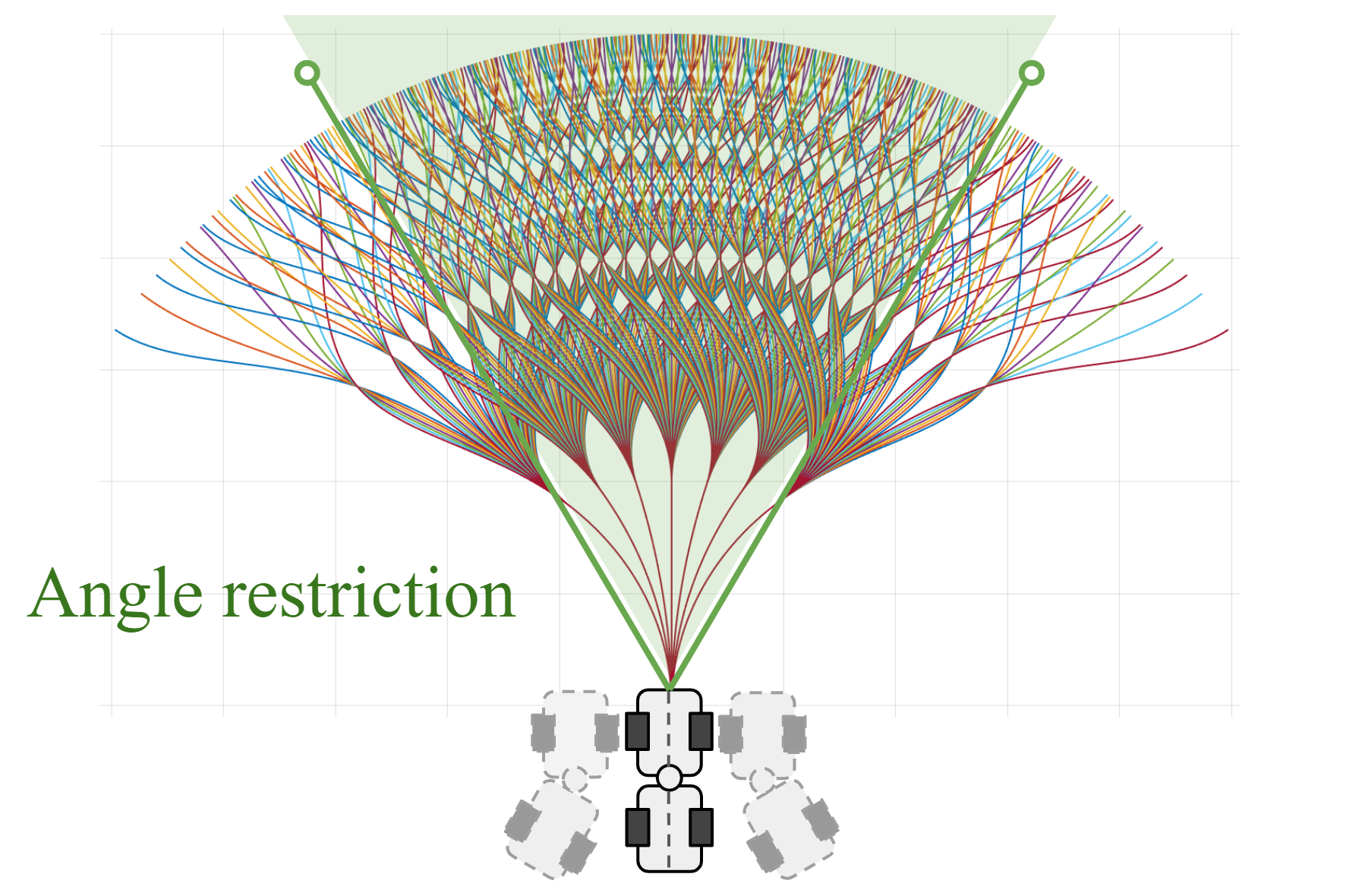}
    \caption{Primitives based on cropped cubic splines}
    \label{fig:previous_primitives}
\end{figure}

We then compare the proposed motion planning pipeline against the baseline method.
The baseline uses a planner adapted from~\cite{Zhang2020} with cropped cubic splines as primitives. These primitives, designed for searching within specified steering angles as depicted in Fig.~\ref{fig:previous_primitives}, enhance the effectiveness of the search process to meet the vehicle's kinematic constraints.
The pure-pursuit (p.p.) controller is used in the baseline to track the path.
In addition, we compare ours with the method that uses our primitives in the local planner, but without using our proposed pose-stabilizing controller for path following.
All approaches utilize a geometric-based terrain analysis module~\cite{cao2022autonomous} for obstacle detection without known prior maps.

To evaluate robustness across various scenarios, we design four environments with varying obstacle numbers \(N_t\) (the number of tree obstacles) and \(N_w\) (the number of wall obstacles).
For each environment, we generated 2 random maps, each spanning \(40m\) by \(40m\) and featuring cylinder and box obstacles to represent trees and walls. 
25 target points were randomly sampled on each map, totaling 200 points across all maps.
As shown in Fig.~\ref{fig:trajs}, the green circles are tree obstacles, gray lines are wall obstacles and the red dots are goals.

The RRT* algorithm, based on Dubins Curve, constrained by the vehicle's steer angle and minimum turning radius, was used to find approximated shortest paths to these points with given prior maps.
During our experiments, we assume after 500,000 iterations, the solution could approximate the optimal path for SAHA.

We assess the algorithm's performance using the Success weighted by Path Length (SPL) metric, as proposed by Anderson and Peter~\cite{anderson2018}.
SPL combines binary success indication with the effectiveness of the path taken relative to the shortest possible path, given by:
\begin{equation}
\text{SPL} = \frac{1}{N} \sum_{i=1}^{N} \frac{S_i l_i}{\max(p_i, l_i)}
\end{equation}
where \( N \) is the total number of test episodes, \( l_i \) is the shortest path distance, \( p_i \) is the path length taken by the agent, and boolean variable \( S_i \) indicates success in episode \( i \).


As shown in Table~\ref{tab:test_results}, methods using our primitives demonstrate a higher success rate and SPL across sparse and dense environments.
This implies planners with our primitives exhibit greater reliability in complex environments.
In addition, compared to the pure-pursuit control law, our control law's advantage grows with the number of obstacles increases. indicating that our control approach can improve the path tracking from primitive sets, especially in environments with dense obstacles.

\subsection{Real-world Experiments}
\begin{figure}[t]
    \centering
    \includegraphics[width=0.48\textwidth]{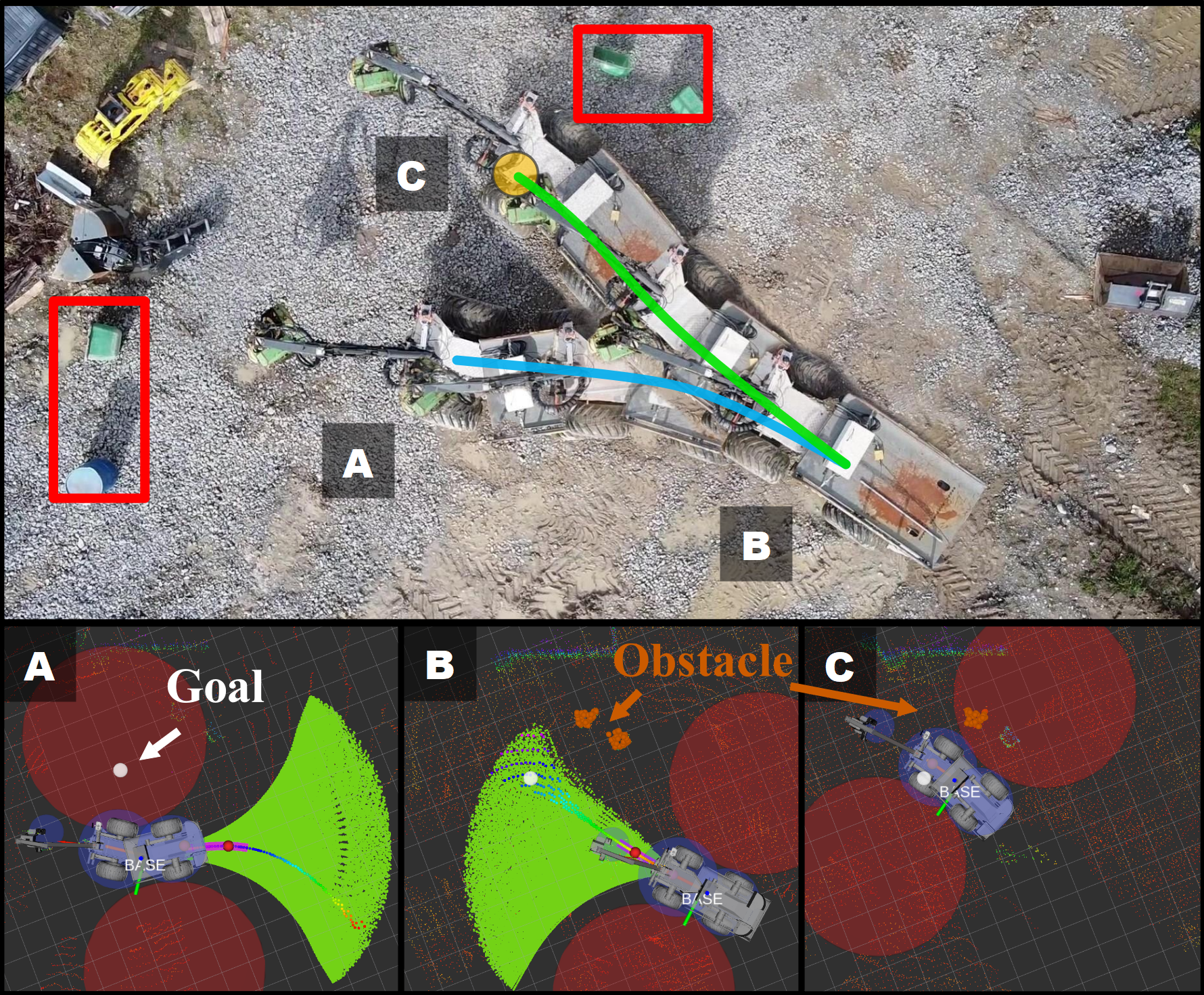}
    \caption{Experiment of bi-directional trajectories. For the image above: obstacles are marked by red boxes. The blue and green curves are backward and forward trajectories. The bottom images demonstrate planning choices in three timestamps: (A) start reversing, (B) start moving forward, and (C) reach goal. The white dot is the goal set in the "Unreachable zone" marked by red areas, and the orange dots are detected obstacles. The green area represents all collision-free paths while the rainbow lines represent chosen paths.}
    \label{fig: zshape}
    \vspace{0.0cm}
\end{figure}

We evaluate the effectiveness of our method on SAHA in an unstructured test field, setting with artificial obstacles. The environment is unknown beforehand, the lidar installed on the right front side of the vehicle (see Fig.~\ref{fig:harveri_real&gazebo}) is used for state estimation and obstacle detection in real-time.
In our experiments, the maximum speed \(v_{max}\) of the vehicle is set to \SI{1.0}{m \per s}.
In the first test, SAHA starts from the entrance of the field and heads to the goal near the edge of the field.
On the way to the goal, there are columnar obstacles and block obstacles of different heights.
To avoid these obstacles, SAHA's final trajectory took on an "S" shape, as shown in Fig.~\ref{fig: opening}.
We also test the planner's ability to choose bi-directional trajectories to reach the goal set in the "Unreachable Zone", which is presented on the two sides of the vehicle due to its limited steering angles.
As shown in Fig.~\ref{fig: zshape}, the vehicle successfully reaches the goal by first reversing and then moving forward.
\vspace{0.5cm}
\section{Conclusion}
\label{sec:conclusion}
This work introduces a motion primitives-based planning strategy and a pose-stabilizing controller designed specifically for CAVs in unstructured environments.
Our method involves a receding horizon planner based on offline-generated motion primitives tailored to the vehicle's kinematic constraints, and a pose-stabilizing controller leveraging the kinematic model to improve path-tracking performance.
We conduct rigorous simulation and real-world testing, including comparative analyses against baseline methods such as cubic spline-based primitives and modified pure-pursuit control, as well as real-world validation on SAHA. Through these evaluations, our method demonstrates significant improvements over previous methods and solid effectiveness in real-world autonomous navigation.
\vspace{1.0cm}


\addtolength{\textheight}{-6.5cm}   








\bibliographystyle{bibliography/IEEEtranN}
\bibliography{bibliography/references}

\end{document}